\documentclass[letterpaper, 10 pt, conference]{ieeeconf}  

\IEEEoverridecommandlockouts                              

\usepackage{amsmath} 
\usepackage{amssymb}  
\usepackage{graphicx} 
\usepackage{tabularx} 
\usepackage{listings}
\usepackage[ruled,linesnumbered]{algorithm2e}

\usepackage{booktabs}
\usepackage{setspace} 
\usepackage{comment}
\usepackage{hyperref} 

\title{\LARGE \bf
ViReSkill: Vision-Grounded Replanning with Skill Memory for LLM-Based Planning in Lifelong Robot Learning
}

\author{Tomoyuki Kagaya*$^{1}$ Subramanian Lakshmi*$^{2}$ Anbang Ye*$^{3}$ Thong Jing Yuan$^{2}$ Jayashree Karlekar$^{2}$ \\
Sugiri Pranata$^{2}$ Natsuki Murakami$^{1}$ Akira Kinose$^{1}$ Yang You$^{3}$
\thanks{*Equal contribution}
\thanks{$^{1}$Panasonic Connect Co., Ltd., Japan, $^{2}$Panasonic R\&D Center, Singapore, $^{3}$National University of Singapore, Singapore. Correspondence to: Tomoyuki Kagaya \texttt{<kagaya.tomoyuki@jp.panasonic.com>}}
}

\begin{document}

\maketitle
\thispagestyle{empty}
\pagestyle{empty}


\begin{abstract}
Robots trained via Reinforcement Learning (RL) or Imitation Learning (IL) often adapt slowly to new tasks, whereas recent Large Language Models (LLMs) and Vision-Language Models (VLMs) promise knowledge-rich planning from minimal data. Deploying LLMs/VLMs for motion planning, however, faces two key obstacles: (i) symbolic plans are rarely grounded in scene geometry and object physics, and (ii) model outputs can vary for identical prompts, undermining execution reliability. 
We propose ViReSkill, a framework that pairs vision-grounded replanning with a skill memory for accumulation and reuse. When a failure occurs, the replanner generates a new action sequence conditioned on the current scene, tailored to the observed state. On success, the executed plan is stored as a reusable skill and replayed in future encounters without additional calls to LLMs/VLMs. This feedback loop enables autonomous continual learning: each attempt immediately expands the skill set and stabilizes subsequent executions.
We evaluate ViReSkill on simulators such as LIBERO and RLBench as well as on a physical robot. Across all settings, it consistently outperforms conventional baselines in task success rate, demonstrating robust sim-to-real generalization.
\end{abstract}


\section{Introduction}

Robots are increasingly expected to master diverse tasks and adapt to dynamic, real-world environments. In such settings, it is essential that agents expand their capabilities throughout their operational lifetime—that is, perform lifelong learning. Traditional motion-generation approaches based on RL and IL typically require numerous demonstrations or extensive trial-and-error, hindering rapid adaptation to new tasks or unseen conditions. Recent advances in LLMs and VLMs suggest that robots can leverage large external knowledge bases to produce plans with minimal task data. Directly applying these models to motion planning, however, raises two critical challenges.

First, a grounding difficulty arises because LLMs and VLMs, while strong in symbolic reasoning, often fail to respect geometric and physical constraints, making it hard to produce action plans that are anchored in the actual environment and the physical properties of target objects.
Second, LLMs' and VLMs' outputs can be unstable owing to hallucination and run-to-run variance, which compromises the reliability of motion plans, especially in physical robotic execution.

We address these issues with a two-pronged framework that couples vision-grounded replanning with skill accumulation and reuse. When a task fails, the robot replans from current observations to generate a physically plausible action sequence. When a task succeeds, the executed plan is stored as a reusable skill in a growing library, enabling subsequent executions without invoking LLMs or VLMs. Through this loop, the robot continually enlarges its skill repertoire and improves online plan quality, thereby adapting to an expanding set of tasks.

We validate the method on LIBERO \cite{liu2023liberobenchmarkingknowledgetransfer}, RLBench \cite{james2019rlbenchrobotlearningbenchmark}, and a real robotic platform. Our experiments show improved task success rates over conventional baselines, confirming both the effectiveness of the approach and its compatibility with lifelong learning across diverse environments.

This paper makes the following contributions:

\begin{itemize}
    \item A vision-grounded replanning strategy that enables failure recovery by generating scene-anchored, feasible motion plans.
    \item A skill-memory execution framework that stores successful plans for zero-inference reuse, expanding the skill set over time and improving stability while reducing LLM/VLM calls.
    \item An empirical evaluation on LIBERO, RLBench, and a physical robot, showing substantial gains in task success rates over a baseline (LIBERO: 45\% → 78\%; RLBench: 47\% → 82\%; real robot: 30\% → 75\%).
\end{itemize}

\begin{figure*}[ht]
    \centering
    \includegraphics[width=\linewidth]{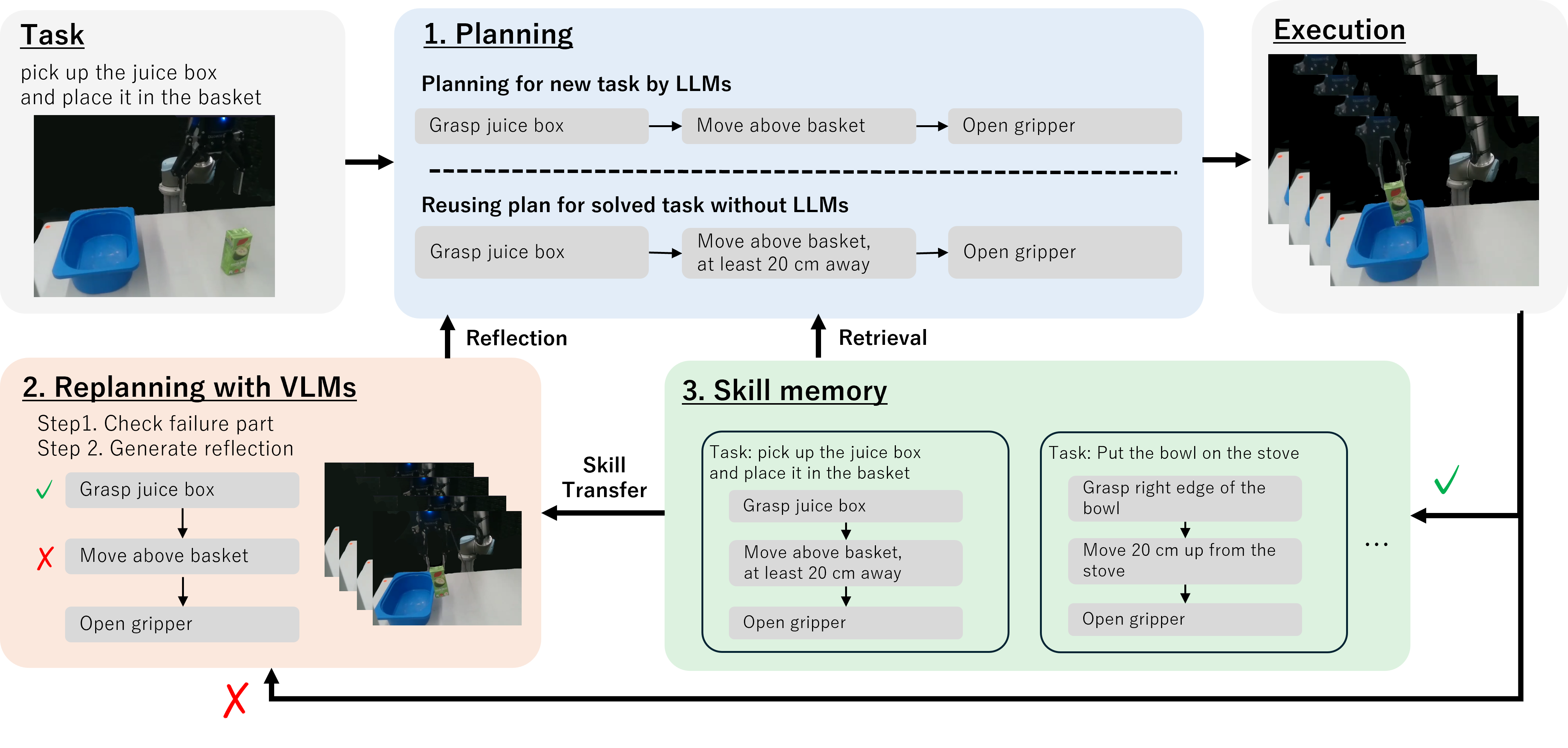}
    \caption{Overview of ViReSkill framework. Given a language-specified manipulation task, an LLM proposes a step-level plan for control. If the task has been solved before, the stored skill is replayed in the \textit{Planning} phase without additional LLM calls, and the robot executes it. On failure, a VLM diagnoses the execution by localizing the first failing step in   the video and explaining the cause; the LLM then repairs the plan in the \textit{Replanning}. On success, the verified plan/code are committed to a task-indexed \textit{Skill memory}. Alternating between vision-grounded replanning on failure and zero-inference replay on success grows the skill library, reduces variance and model calls, and improves success rate over time.}
    \label{fig:vireskill-pipeline}
\end{figure*}

\section{Related Work}
\label{sec:related_work}

\subsection{LLM/VLM-based Planning and Manipulation}

SayCan \cite{ahn2022icanisay} grounds language plans in executability by filtering LLM proposals with affordances from RL policies. VoxPoser \cite{voxposer} converts language into 3D affordance/constraint maps to produce scene-consistent, zero-shot manipulation. ReKep \cite{huang2024rekepspatiotemporalreasoningrelational} encodes relations among task-relevant 3D keypoints as explicit costs and solves a vision-conditioned optimization for feasible trajectories. These approaches enable knowledge-driven planning but stop short of a closed loop that replans from live observations and systematically reuses successful executions.
ViReSkill differs in that it closes this loop by combining vision-grounded replanning with direct reuse of successful plans via skill memory.

\subsection{Replanning and Grounded Feasibility}

Methods targeting the plan–execution gap include Reflexion (self-reflection memory for trial-over-trial improvement) \cite{shinn2023reflexionlanguageagentsverbal}, REFLECT (LLM-guided failure analysis and fixes) \cite{liu2023reflectsummarizingrobotexperiences}, DoReMi (LLM constraints with VLM violation detection) \cite{guo2024doremigroundinglanguagemodel}, LLM+A (prompted goal-conditioned affordances) \cite{cheng2024empoweringlargelanguagemodels}, and Phoenix (self-reflection mapped to diffusion-policy corrections) \cite{xia2025phoenixmotionbasedselfreflectionframework}. They improve diagnosis and online correction, yet typically do not unify vision-grounded replanning with immediate consolidation of successes into reusable skills.

\subsection{Skill Libraries and Lifelong/Continual Learning}

RoboCat \cite{bousmalis2023robocatselfimprovinggeneralistagent} leverages cross-embodiment experience and self-generated data for general manipulation; RoboMatrix \cite{mao2025robomatrixskillcentrichierarchicalframework} composes meta-skills to solve unseen tasks; LRLL \cite{tziafas2024lifelongrobotlibrarylearning} grows a library by distilling successful trajectories. ViReSkill builds on this line by coupling vision-grounded replanning (on failure) with skill memory for zero-inference replay (on success) in one loop, reducing model calls while stabilizing execution and expanding capabilities over time.


\section{Proposed Method}
\label{sec:method}

We introduce the ViReSkill framework to address the challenge of lifelong skill learning in LLM-based robot manipulation. In this section, we first provide the problem formulation \hyperref[subsec:problem_formulation]{(§\ref*{subsec:problem_formulation})}, which outlines the setting and objectives of our lifelong learning approach. Based on this formulation, we describe how LLMs are leveraged for task planning \hyperref[subsec:llm_planning]{(§\ref*{subsec:llm_planning})}, enabling flexible and generalizable behavior composition. We then show how lifelong learning can be achieved through our skill-based execution framework \hyperref[subsec:skill_based_framework]{(§\ref*{subsec:skill_based_framework})} and the vision-conditioned replanning method \hyperref[subsec:vision_based_replanning]{(§\ref*{subsec:vision_based_replanning})} for recovering from failure and exploring novel policies.

\subsection{Problem Setup and Overview}
\label{subsec:problem_formulation}

We study a lifelong learning problem in which a robot repeatedly solves manipulation tasks through trial-and-error and progressively improves its accumulated skills within the ViReSkill framework. The setup is as follows:

\begin{itemize}
\item \textbf{Tasks}: Let $\mathcal{T}_i$ be the $i$-th manipulation task in the training set, described by a free-form language input $\mathcal{D}_i$. The full training set is $\{\mathcal{T}_i\}_{i=1}^M$.
\item \textbf{Skill Repertoire}: For each task $\mathcal{T}_i$, we define its associated skill as $\pi_i \equiv \mathcal{P}(\mathcal{T}_i)$, which consists of a high-level plan and low-level control code. The full repertoire is denoted by $\mathcal{P} = \{\pi_i\}_{i=1}^M$.
\item \textbf{Rollout and Success Rate}: The executor $\mathcal{C}$ (based on VoxPoser \cite{voxposer}) takes as input a skill $\pi_i$, task $\mathcal{T}_i$, and stochastic factors $\xi$ (e.g., object pose variation, non-deterministic dynamics), and returns a motion trajectory $\tau$. Each rollout is evaluated by a binary success function $r(\tau, \mathcal{T}_i) \in \{0, 1\}$.
\end{itemize}

The success rate (SR) of skill $\pi_i$ on task $\mathcal{T}_i$ over $N$ independent rollouts is:
\begin{align}
    \mathrm{SR}(\pi_i, \mathcal{T}_i)
    &= \frac{1}{N} \sum_{n=1}^{N} r\!\left(\tau(\xi_n), \mathcal{T}_i\right),
    \label{formula:success_rate}\\
    \tau(\xi) &= \mathcal{C}(\pi_i, \mathcal{T}_i, \xi),
    \label{formula:rollout}
\end{align}
where $\xi_n$ denotes the stochastic input for the $n$-th rollout.

Our objective is to update the skill repertoire over iterations so as to maximize the average success rate across all training tasks:
\begin{align}
    \mathcal{P}^\ast
    = \arg\max_{\mathcal{P}} \;
    \frac{1}{M} \sum_{i=1}^{M} \mathrm{SR}(\pi_i, \mathcal{T}_i).
\end{align}

Algorithm \ref{algo:vireskill_framework_train} outlines the training process, while Algorithm \ref{algo:vireskill_framework_eval} details the evaluation procedure.

\begin{algorithm}[t]
\caption{Training Stage of The ViReSkill Framework}\label{algo:vireskill_framework_train}
\KwData{Training tasks $\{\mathcal{T}\}_{i=1}^{M}$, corresponding task description $\{\mathcal{D}\}_{i=1}^{M}$, LLM-based agent $\mathcal{C}$}
\KwResult{Skill memory $\{\mathcal{P}_k\}_{k=1}^{10}$ at each iteration.}
$\mathcal{P}_1=\{\mathcal{P}_1(\mathcal{T}_i)\leftarrow \mathtt{null}\}_{i=1}^{M}$\\
$c \leftarrow \mathtt{null}$\tcp*[f]{Initialize control code}\\
\For(\tcp*[f]{For each training round}){$j\leftarrow 1$ \KwTo $2$}{
    \For(\tcp*[f]{For each task}){$i\leftarrow 1$ \KwTo $M$}{
        $iter \leftarrow 1 + 5\times(j-1)$\\
        \emph{Initialize control code from skill memory if skill exists}\\
        $\mathcal{P}_{iter}(\mathcal{T}_i)\leftarrow \mathcal{P}_{iter - 1}(\mathcal{T}_i)$\\
        \If{$\mathcal{P}_{iter}(\mathcal{T}_i) \neq \mathtt{null}$}{
            $c \leftarrow \mathcal{P}_{iter}(\mathcal{T}_i)$\tcp*[f]{reuse skill}\\
        }
        \Else{$c \leftarrow \mathcal{C}.planning(\mathcal{D}_i)$\tcp*[f]{use control code generated by agent}\\}
        \For(\tcp*[f]{For each training iteration}){$k\leftarrow 1$ \KwTo $5$}{
            $\tau \leftarrow \mathcal{C}.verify(c, \mathcal{T}_i)$\\
            $r, \mathcal{O} \leftarrow r(\tau, \mathcal{T}_i)$\tcp*[f]{collect rollout success status and observations}\\
            \If(\tcp*[f]{replan if failed}){$r \neq 1$}{
                $c \leftarrow \mathcal{C}.replanning(c, \mathcal{D}_i, \mathcal{O}, \mathcal{P}_{iter})$\tcp*[f]{replanning with skill}\\
            }
            \Else{$\mathcal{P}_{iter}\leftarrow c$\tcp*[f]{Update skill memory}}
            $iter \leftarrow iter + 1$
        }
    }
}
\end{algorithm}

\begin{algorithm}
\caption{Evaluation Stage of The ViReSkill Framework}\label{algo:vireskill_framework_eval}
\KwData{Training tasks $\{\mathcal{T}\}_{i=1}^{M}$, Skill memory $\{\mathcal{P}_k\}_{k=1}^{10}$ at each iteration, LLM-based agent $\mathcal{C}$}
\KwResult{Success rate for each task at each iteration $\{success\_rate_{ij}\}_{i=1,j=1}^{i=M,j=10}$}\
\For(\tcp*[f]{For each task}){$i\leftarrow 1$ \KwTo $M$}{
    \For(\tcp*[f]{For each skill memory (2 rounds 5 iterations per round) to be evaluated}){$j\leftarrow 1$ \KwTo $10$}{
        $success\_count\leftarrow 0$\\
        \For(\tcp*[f]{For each evaluation trial}){$k\leftarrow 1$ \KwTo $5$}{
            \If{$\mathcal{P}_{j}(\mathcal{T}_i) \neq \mathtt{null}$}{
                $c\leftarrow \mathcal{P}_{j}(\mathcal{T}_i)$\\
                $\tau \leftarrow \mathcal{C}.verify(c, \mathcal{T}_i)$\\
                $r, \mathcal{O} \leftarrow r(\tau, \mathcal{T}_i)$\tcp*[f]{collect rollout success status and observations}\\
                $success\_count \leftarrow success\_count + r$\\
            }
        }
        $success\_rate_{ij}\leftarrow success\_count / 5$
    }
}
\end{algorithm}

\subsection{Planning by LLMs}
\label{subsec:llm_planning}

    We use LLMs to generate Python code that is executed by an interpreter to decompose the language instruction $\mathcal{D}_i$ of task $\mathcal{T}_i$ into subtasks, invoke perception APIs, and generate robot trajectories $\tau$.  
    Let $d_j$ denote the $j$-th subtask instruction derived from $\mathcal{D}_i$, such that the sequence of subtask instructions is $(d_1, d_2, \dots, d_n)$.  
    Our planning pipeline follows the design from VoxPoser \cite{voxposer}, and consists of three steps involving LLM-based code generation:
    
    Step 1: The Planner is responsible for decomposing the natural language instruction $\mathcal{D}_i$ (e.g., ``press the light switch'') into a sequence of structured subtask instructions:
    \begin{equation}
        \label{eqn:planner_d}
        Planner(\mathcal{D}_i) = (d_1, d_2, \dots, d_n)
    \end{equation}
    Each $d_j$ corresponds to an interpretable subtask (e.g., ``grasp the button'', ``move to the center of the button'') for task $\mathcal{T}_i$.

    Step 2: The Composer takes as input a subtask instruction $d_j$ and constructs a set of Language Model Programs (LMPs), each responsible for a distinct function such as object detection, affordance prediction, or motion planning:
    \begin{equation}
        \label{eqn:composer_d}
        Composer(d_j) = \left( LMP_1(d_j), LMP_2(d_j), \dots, LMP_k(d_j) \right)
    \end{equation}

    Step 3: The Low-Level LMPs returned by the Composer are executed via the executor $\mathcal{C}$ to produce a sub-trajectory $\tau_j$:
    \begin{equation}
        \label{eqn:lmp_exec_d}
        \tau_j = \mathcal{C}(Composer(d_j), \mathcal{T}_i, \xi)
    \end{equation}
    
    We define the overall skill $\pi_i$ as the sequence of all LMP modules returned by the Composer for each subtask instruction $d_j$:
    \begin{equation}
        \label{eqn:pi_def_composer}
        \pi_i = \left( Composer(d_1), Composer(d_2), \dots, Composer(d_n) \right)
    \end{equation}
    
    The full robot trajectory $\tau$ for task $\mathcal{T}_i$ is formed by concatenating all sub-trajectories:
    \begin{equation}
        \label{eqn:traj_d}
        \tau = (\tau_1, \tau_2, \dots, \tau_n)
    \end{equation}

\begin{figure*}[ht]
\centering
\includegraphics[width=\linewidth]{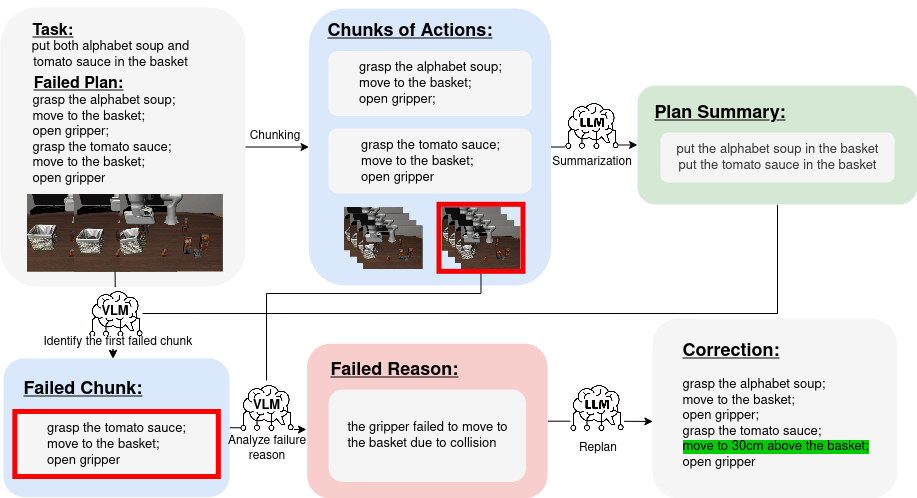} 
\caption{Overview of the proposed hierarchical video-based reflection mechanism. The textual plan and execution video are segmented into logical chunks of actions (via “open gripper” as a heuristic). Each chunk is summarized into a high-level natural language description, aligned with the corresponding video segment, and evaluated by a VLM to localize failures. The VLM then provides explicit failure reasons, which are used together with the original plan to generate corrections with the LLM.}
\label{fig:visual_reflection}
\end{figure*}

\subsection{Skill Memory for Lifelong Robot Learning}
\label{subsec:skill_based_framework}

We implemented the ViReSkill to enable lifelong skill learning in LLM-based robotic manipulation. The system maintains a task-specific skill memory comprising both high-level and low-level control code. This memory is updated iteratively through replanning upon failure.

At the beginning of training, the skill memory is empty, meaning the robot has no prior successful experiences to reference. To address this cold start condition, the agent relies solely on LLM-based planning to generate an initial high-level and low-level control code from the task description. Once a trial for task $\mathcal{T}_i$ succeeds, the corresponding control code is used to update the task-specific memory $\mathcal{P}(\mathcal{T}_i)$. Due to stochastic variations in rollouts, a skill that has succeeded once may fail in subsequent trials of the same task, resulting in a non-saturated success rate and leaving room for iterative refinement.

A key challenge in policy optimization for robotic manipulation—whether in trainable or agent-based approaches—is the exploration-exploitation dilemma. To balance exploitation and exploration, we design a training stage that consists of 10 iterations, organized into 2 rounds of 5 iterations each.

At the start of each round, if a skill exists in memory for the given task, the LLM-based agent initializes its control code with the stored version to maximize exploitation. Otherwise, the agent generates a new control code based on the task description. During the five iterations of a round, the agent collects success indicators and visual evidence (video recording of the scene collected by the front camera) from the environment. If an iteration succeeds, the control code remains unchanged. Upon failure, the agent explores new high-level plans through engaging in vision-based replanning—a reflection mechanism that revises the failed high-level plan given visual evidence (detailed in Section \ref{subsec:vision_based_replanning}). Since each round has five iterations, the agent can perform up to four replanning attempts per round. We observe that additional replanning beyond four attempts yields negligible improvement; thus, it is more effective to begin a new round, reinitializing the control code from the latest successful version stored in memory.

For evaluating lifelong learning performance, we saved the skill memory after each iteration and used the saved memory to compute the average success rate using (\ref{formula:success_rate}). Specifically, for each skill iteration, we compute the mean success rate across five rollouts ($N=5$).

\subsection{Vision-Grounded Replanning with Skill Transfer}
\label{subsec:vision_based_replanning}

While LLMs have demonstrated strong planning capabilities, their outputs often overlook physical constraints and lack grounding in real-world dynamics. To address these limitations, we adopt a self-learning scheme that combines current observations with past successful experiences, leveraging methods that support retrieval-augmented planning \cite{kagaya2024rapretrievalaugmentedplanningcontextual} and vision-based replanning \cite{liu2023reflectsummarizingrobotexperiences}.
For replanning, we retrieve a mixed set of examples based on (\ref{formula:retrieval}): half selected by task similarity and half selected by plan-code line-level similarity between the failed plan and the stored plans in the skill memory.

Specifically, let $\mathcal{D}$ denote the task description, and $\{d_1, d_2, \dots, d_n\}$ be the sequence of subtask instructions generated from $\mathcal{D}$ by the planner, as defined in Formula \ref{eqn:planner_d}. Each $d_i$ corresponds to a high-level code line in the failed plan.
For task similarity-based retrieval, we embed the entire task description $\mathcal{D}$ and select top examples based on cosine similarity with stored tasks $\mathcal{D}’$.
For code-level similarity, each subtask $d_i$ is embedded and matched to the most similar past subtask $d’_j$, and examples are ranked by the average of these maximum similarities.
To ensure relevance, we further filter the candidates retrieved by task similarity and discard candidates with task similarity scores below a threshold (e.g., $0.5$). The remaining retrieved examples are then injected into the code generation or re-planning prompt to facilitate skill transfer and promote effective exploitation.

\begin{equation}
\begin{aligned}
\mathcal{R}(\mathcal{D}) = 
&\;\underset{>0.5}{\mathrm{filter}}\big(\underset{(\mathcal{D}') \in \mathcal{P}}{\mathrm{Top}\text{-}\tfrac{k}{2}}
   \cos\big(\mathrm{emb}(\mathcal{D}),\; \mathrm{emb}(\mathcal{D}') \big)\big) \; \cup\\
&
  \;\underset{(\mathcal{D}') \in \mathcal{P}}{\mathrm{Top}\text{-}\tfrac{k}{2}}
   \; \frac{1}{n} \sum_{i=1}^{n} \max_{j} 
   \; \cos \big( \mathrm{emb}(d_i), \mathrm{emb}(d'_j) \big),
\end{aligned}
\label{formula:retrieval}
\end{equation}

Reflection with visual input presents several challenges, including the need to reason over long-horizon tasks, the visual language model's limited ability to understand complex plan code, and the constraint on the number of visual frames that can be provided as evidence.
To address those challenges, we introduce a hierarchical video-based reflection mechanism that localizes and analyzes failures through chunked video-based reflection as illustrated in Fig.~\ref{fig:visual_reflection}.

First, we segment both the high-level plan and the execution video into logical chunks using a simple heuristic: splitting at each “open gripper” action. Each chunk is then summarized separately and concatenated to form a concise natural language description of the plan, which helps reduce confusion for the VLM. The VLM compares the summarized plan with the raw video to identify the first failure chunk. Once a failed chunk is identified, the VLM analyzes both the chunked plan code and the corresponding video chunk to provide a grounded explanation of the failure. To handle plan-level logical errors \cite{liu2023reflectsummarizingrobotexperiences}—cases where every individual step is executed correctly, but the overall task fails due to flawed logic (e.g., grasping a new object without first releasing the one already held)—the system can explicitly indicate that all actions were executed successfully, which triggers plan-level reflection on the high-level plan with a LLM for checking flawed logic. 

Finally, the LLM generates a revised plan based on both the detected failure reason and the original plan code. Separate replanning prompts are used depending on whether the error is due to an execution failure or a logical error. Inspired by the DROC system \cite{DROC}, we ground the replanning process on properties of related objects, for instance, determining a proper offset and obstacle avoidance strategy based on the scale of related objects.


\section{Experiments}
\label{sec:experiments}

\subsection{Setup: Benchmarks, Real Robot, and Metrics}

To evaluate the effectiveness of the proposed ViReSkill framework, we conduct extensive experiments across both simulation and real-world settings. We benchmark ViReSkill along with several baselines on two robot manipulation benchmarks that operate in simulation environments: LIBERO and RLBench. We also validate the framework under a real-world setting using a physical UR5 robotic platform, thereby demonstrating the transferability of learned skills from simulation to practice. We select simulation benchmarks based on three criteria: appropriate task difficulty, widespread adoption in robot manipulation research, and their representativeness of real-world manipulation scenarios. 

For baseline setups, we compare ViReSkill with the other four baseline methods, including VoxPoser \cite{voxposer}, and three replanning methods: Retry, Reflexion \cite{shinn2023reflexionlanguageagentsverbal}, and REFLECT \cite{liu2023reflectsummarizingrobotexperiences}. A summary of these baselines is provided below:
\begin{itemize}
    \item \textbf{VoxPoser}: The base code generation pipeline without replanning. A single control plan is generated per task and executed to compute success rate.
    \item \textbf{Retry}: A naive baseline that regenerates a new control plan from scratch using the LLM after each failure, without referencing prior attempts or stored context.
    \item \textbf{Reflexion}: A vision-based reflection method that analyzes the final frame of a failed attempt together with the failed plan to infer the failure reason and propose a corrective plan. Unlike our ViReSkill framework, Reflexion does not incorporate hierarchical reflection or in-context learning from learned skill memory.
    \item \textbf{REFLECT}: A LLM-based replanning method that hierarchically summarizes multisensory observation, analyzes failure reasons hierarchically, and generates a new plan.
\end{itemize}
We use all four baselines for the LIBERO experiments, while VoxPoser, Retry, and Reflexion are used for the RLBench experiments. For the real-robot experiments, we compare ViReSkill against the VoxPoser baseline.

We use the success rate, the percentage of successful trial during evaluation, given by (\ref{formula:success_rate}), as the metric to measure the life-long learning performance of all baselines across different benchmarks. For the VoxPoser experiments, since it does not involve replanning, we calculate the success rate over 5 trials for each task. For ViReSkill, Retry, Reflexion, and REFLECT, we calculate the success rate for each training iterations using the saved memory snapshots. For each task, if a solution exists in the memory, we perform 5 trials using the memory to calculate the success rate, otherwise, the success rate of the task is zero.

\subsection{LIBERO Benchmark Results}

LIBERO \cite{liu2023liberobenchmarkingknowledgetransfer} is a robot manipulation benchmark designed for evaluating the knowledge transfer in lifelong robot learning. It groups evaluation tasks into 5 subsets: LIBERO-Object, LIBERO-Spatial, LIBERO-Goal, LIBERO-10, and LIBERO-90 with each subset containing tasks of comparable difficulty. For our experiments, we evaluate on the LIBERO-Object, LIBERO-Spatial, LIBERO-Goal, and LIBERO-10 subsets, each comprising 10 distinct tasks.

For ViReSkill, Reflexion, and REFLECT, we use visual observations captured from the front camera of the LIBERO simulation environment at a resolution of $256 \times 256$ pixels. In ViReSkill, input videos are uniformly downsampled to 30 frames. For REFLECT, RGB-D videos of the same resolution are collected to reconstruct dynamic 3D scenes, and additional “grasped object” states—computed using heuristics—are provided to the REFLECT system.

Across all baselines, we employ GPT-4o-mini as the VLM for analyzing video or image-based failure cases, and GPT-4.1-mini as the language model (LLM) for code generation and replanning.

\begin{table}[t]
\caption{Last iteration performance comparison for LIBERO benchmark.}
\centering
\small
\setlength{\tabcolsep}{4pt} 

\begin{tabular}{rccccc}
\hline
Benchmark & VoxPoser & Retry & Reflexion & REFLECT & ViReSkill \\
\hline
Object  & 0.42 & 0.46 & \textbf{0.94} & 0.68 & \textbf{0.94} \\
Spatial  & 0.52 & 0.52 & 0.72 & 0.78 & \textbf{0.80} \\
Goal  & 0.54 & 0.48 & \textbf{0.70} & \textbf{0.70} & \textbf{0.70} \\
10  & 0.32 & 0.36 & 0.46 & 0.48 & \textbf{0.66} \\
\hline
Average  & 0.45 & 0.46 & 0.71 & 0.66 & \textbf{0.78} \\
\hline
\end{tabular}
\label{tab:last_iteration_comparison}
\end{table}

\begin{figure}[ht]
\centering
\includegraphics[width=\linewidth]{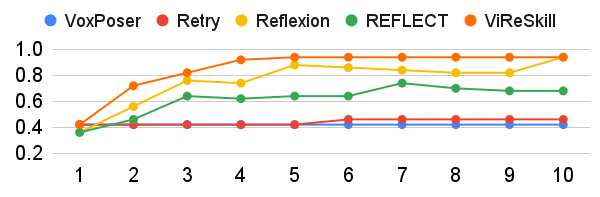}
\centering
\includegraphics[width=\linewidth]{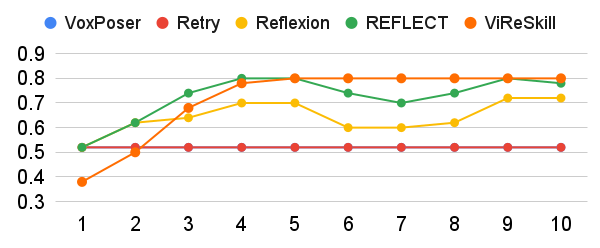}
\centering
\includegraphics[width=\linewidth]{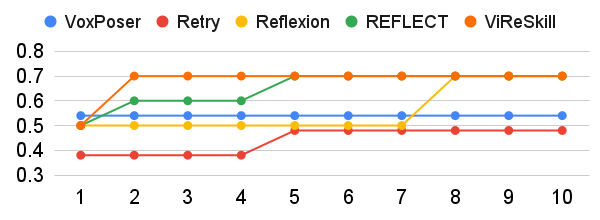}
\centering
\includegraphics[width=\linewidth]{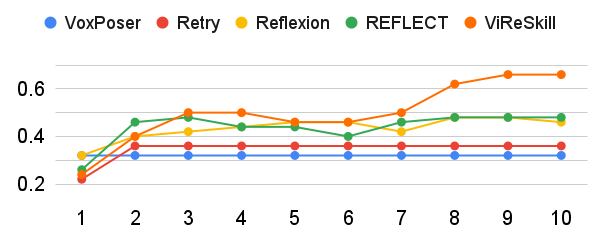}

\centering
\caption{Success rate of methods on LIBERO benchmark across iterations. Top to bottom: LIBERO-Object, LIBERO-Spatial, LIBERO-Goal, LIBERO-10. The horizontal axis is iterations, the vertical axis is success rate (ranging from 0 to 1).}
\label{fig:libero_line_chart}
\end{figure}

Table~\ref{tab:last_iteration_comparison} shows the mean success rate of all baselines at the last training iteration, averaged over all 10 tasks on each LIBERO subset.  We visualize the change of success rate over iterations in Fig.~\ref{fig:libero_line_chart}.
Across all benchmarks, our ViReSkill framework consistently outperforms the baseline methods. Moreover, ViReSkill demonstrates stable and substantial performance gains over successive iterations, underscoring the effectiveness of the proposed visually grounded reflection mechanism. Notably, while the video-grounded replanning approach achieves performance comparable to the image-grounded self-reflection method on relatively simple tasks (LIBERO-Object, LIBERO-Spatial, LIBERO-Goal), it significantly surpasses other methods on the more challenging LIBERO-10 subset. This subset involves composite tasks (e.g., placing both the mokapot and the cup on the stove), which demand reasoning over temporal dependencies. These results highlight the importance of leveraging temporal observations for effective replanning in robotic manipulation tasks, particularly in complex, multi-stage scenarios.

\begin{table}[t]
\caption{Performance comparison for RLBench.}
\centering
\small
\setlength{\tabcolsep}{4pt}
\begin{tabular}{rccccc}
\hline
Task & VoxPoser & Retry & Reflexion & ViReSkill \\
\hline
TakeLidOff  & 0.00 & 0.00 & \textbf{1.00} & \textbf{1.00} \\
CloseDrawer  & 0.40 & \textbf{1.00} & 0.60 & \textbf{1.00} \\
BasketballInHoop  & 0.20 & 0.20 & 0.40 & \textbf{0.60} \\
BeatTheBuzz  & \textbf{0.60} & \textbf{0.60} & \textbf{0.60} & \textbf{0.60} \\
PutRubbishInBin  & 0.60 & \textbf{0.80} & \textbf{0.80} & \textbf{0.80} \\
LampOff  & \textbf{1.00} & 0.80 & 0.80 & 0.60 \\
OpenWineBottle  & 0.60 & \textbf{1.00} & 0.80 & 0.80 \\
PushButton  & 0.00 & 0.00 & \textbf{1.00} & \textbf{1.00} \\
TakeUmbrellaOut & 0.80 & 0.80 & \textbf{1.00} & \textbf{1.00} \\
\hline
Average & 0.47 & 0.58 & 0.78 & \textbf{0.82} \\
\hline
\end{tabular}
\label{tab:rlbench_result}
\end{table}

\subsection{RLBench Benchmark Results}

Beyond the LIBERO benchmark, we further evaluate ViReSkill on RLBench, which consists of 100 hand-designed robotic manipulation tasks in a simulated tabletop environment. For our experiments, we select 9 tasks that cover a range of difficulties and task types. Visual observations are obtained from the front camera at a resolution of $128 \times 128$ pixels, and the experimental setup closely follows the one used in the LIBERO experiments. Although RLBench offers multiple instructions for each task, this evaluation used a single, fixed instruction.

Table~\ref{tab:rlbench_result} shows the comparison between the ViReSkill framework and baseline methods.
ViReSkill achieved the highest success rate on RLBench (82\%), highlighting the effectiveness of combining replanning with skill memory. Compared to static planning in VoxPoser, ViReSkill improves performance through memory-based execution. Unlike Retry’s blind regeneration, it performs targeted replanning based on failure analysis. Although Reflexion also benefits from failure-aware replanning, ViReSkill slightly outperforms it by additionally reusing verified skills. These trends resemble those on simpler LIBERO subsets, suggesting structural similarities.

\begin{figure}[ht]
    \centering
    \includegraphics[width=\linewidth]{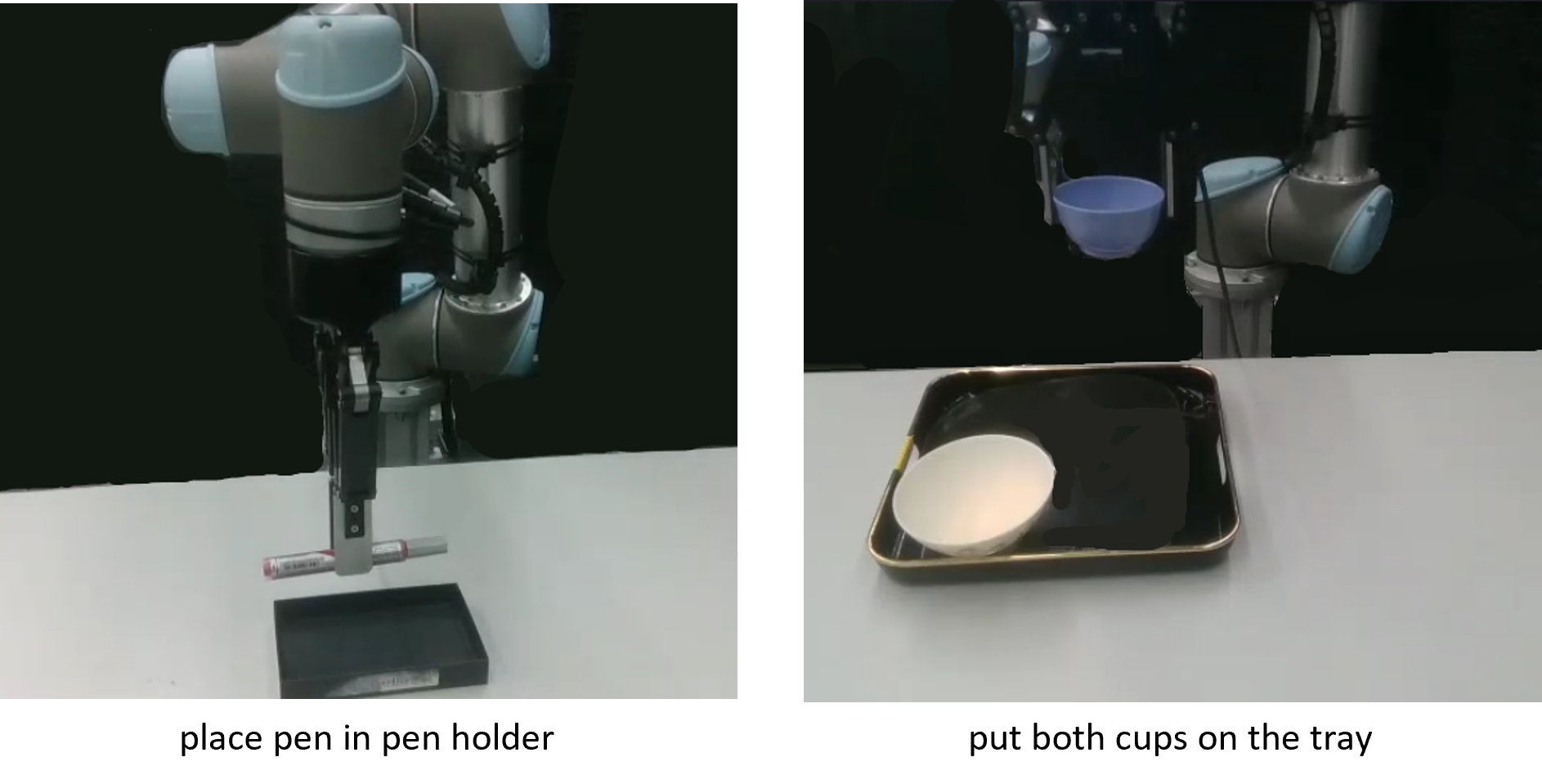}
    \vskip -0.1in
    \caption{Demonstration of UR5 executing diverse real-world tasks using ViReSkill}
    \label{fig:ur5_tasks}
\end{figure}

\begin{table*}[h]
\centering
\caption{A comparison between VoxPoser and ViReSkill in planning the placement of two cups on a tray}
\label{tab:plan_comparison}
\begin{tabular}{@{}p{0.46\textwidth}@{\hspace{0.01\textwidth}}p{0.46\textwidth}@{}}
\toprule
\textbf{VoxPoser} & \textbf{ViReSkill} \\
\midrule

\begin{minipage}[t]{\linewidth}
\scriptsize
\setstretch{1}
\begin{flushleft}
\texttt{composer("grasp the blue cup")} \\
\texttt{composer("back to default pose")} \\
\texttt{composer("move to 5cm on top of the black tray")} \\
\texttt{composer("open gripper")} \\
\texttt{composer("back to default pose")} \\
\texttt{composer("grasp the white cup")} \\
\texttt{composer("back to default pose")} \\
\texttt{composer("move to 5cm on top of the black tray")} \\
\texttt{composer("open gripper")}
\end{flushleft}
\end{minipage}
&
\begin{minipage}[t]{\linewidth}
\scriptsize
\setstretch{1}
\begin{flushleft}
\texttt{composer("back to default pose")} \\
\texttt{composer("grasp the blue cup")} \\
\texttt{composer("move to 5cm on top of the front-left part of the black tray")} \\
\texttt{composer("open gripper")} \\
\texttt{composer("back to default pose")} \\
\texttt{composer("grasp the white cup")} \\
\texttt{composer("move to 5cm on top of the back-right part of the black tray")} \\
\texttt{composer("open gripper")} \\
\texttt{composer("back to default pose")}
\end{flushleft}
\end{minipage} \\

\bottomrule
\end{tabular}
\end{table*}

    \begin{figure*}[ht]
        \centering
        \includegraphics[width=\linewidth]{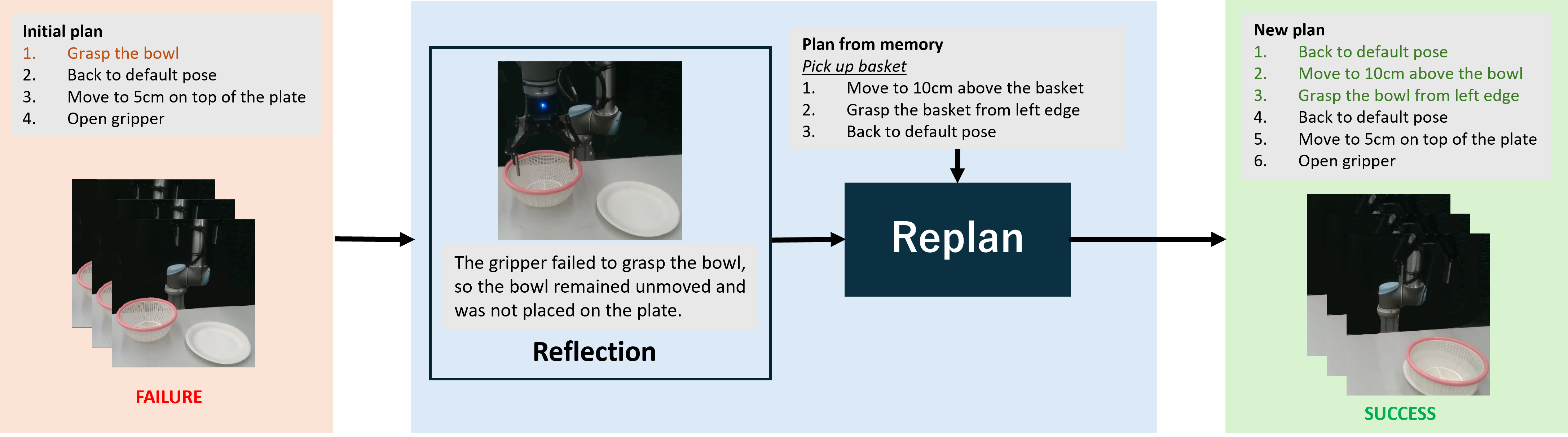}
        \vskip -0.1in
        \caption{Demonstration of Task: Pick up bowl and place it on the plate using ViReSkill on UR5}
        \label{fig:ur5_bowl_tasks}
    \end{figure*}

\subsection{Real-robot Evaluation Results}

We conducted our experiments on a UR5 robot equipped with a Robotiq 2F gripper and an Intel RealSense \cite{intelRealsense2020} RGB-D camera. LangSAM \cite{medeiros2025langsegmentanything} was employed for object segmentation, while GPT-4.1-mini handled vision-language tasks such as video understanding and plan generation. Sentence Transformer \cite{reimers2019sentencebert} model was used to retrieve relevant past experiences from the memory constructed from Libero experiments. We evaluated four representative tasks: (i) placing an object in a basket, (ii) placing a bowl on a plate, (iii) placing two cups on a tray, and (iv) placing a pen in a holder. Examples of these tasks executed on the UR5 robot are shown in Fig.~\ref{fig:ur5_tasks}. Each task was executed five times using both VoxPoser and our proposed method for comparison. Table~\ref{tab:ur5_task_comparison} shows our video-based replanning method significantly outperforms VoxPoser, achieving a 75\% vs. 30\% success rate. Our method detects failed steps—such as missed grasps or poor object handling—and dynamically adjusts the plan.
For instance, in the task of picking and placing an object into a basket, failures sometimes arise during the grasping or placing stages. Our system detects when the grasp is unsuccessful and adjusts the plan to improve the grip, resulting in a more reliable manipulation. Similarly, when dealing with larger objects that cannot be effectively grasped at their center, the algorithm adapts by replanning to grasp the object by its edges as shown in Fig.~\ref{fig:ur5_bowl_tasks}, ensuring the task can still be completed successfully.
In placement tasks like putting cups on a tray, a straightforward plan might place all cups centrally. However, our approach recognizes when such placement is unstable or incorrect, and modifies the plan to distribute the cups across different positions on the tray. See Table~\ref{tab:plan_comparison} for the plans generated by both methods for this task. Likewise, when placing a pen in a holder, where orientation is critical but not explicitly indicated in the initial plan, our method adjusts the orientation to correctly position the pen in the holder.
Through this experience-driven, feedback-informed replanning, our algorithm achieves a greater level of robustness and adaptability during real-world task execution, complementing the strong baseline plans generated by VoxPoser with enhanced flexibility and precision.

\begin{table}[t]
    \caption{Task success rates on the UR5 robot comparing VoxPoser and our ViReSkill method (successes out of 5 trials).}
    \centering
    \small
    \setlength{\tabcolsep}{4pt}
    \begin{tabular}{lcc}
    \hline
    Tasks & VoxPoser & ViReSkill \\
    \hline
    Place juice in basket     & 2/5 & 4/5 \\
    Place bowl on plate        & 2/5 & 4/5 \\
    Place both cups on tray    & 0/5 & 3/5 \\
    Place pen in holder        & 3/5 & 4/5 \\
    \hline
    Overall Success Rate & 30\% & 75\% \\
    \hline
    \end{tabular}
    \label{tab:ur5_task_comparison}
\end{table}

\subsection{Ablation Study}

An important element of the ViReSkill framework is its ability to reuse skills from previously solved tasks to support new or related ones—enabling effective lifelong learning.
To evaluate this, we compare the full ViReSkill against a variant that disables skill transfer during replanning.
The results (Table~\ref{tab:ablation_knowledge_transfer}) show that while both methods perform similarly on simpler LIBERO subsets, ViReSkill significantly outperforms the ViReSkill without skill transfer on LIBERO‑10 (0.66 vs. 0.50).
These results support that skill transfer is a key driver of lifelong learning in ViReSkill, enabling generalization and faster adaptation to complex, multi-stage tasks.

\begin{table}[t]
\centering
\small
\setlength{\tabcolsep}{4pt}
\caption{Performance comparison with and without skill transfer for LIBERO.}
\label{tab:ablation_knowledge_transfer}
\begin{tabular}{lcc}
\hline
  & ViReSkill w/o skill transfer & ViReSkill \\
\hline
LIBERO-Object & \textbf{0.94} & \textbf{0.94} \\
LIBERO-Spatial & 0.78 & \textbf{0.80}\\
LIBERO-Goal & \textbf{0.70} & \textbf{0.70} \\   
LIBERO-10 & 0.50 & \textbf{0.66} \\
\hline
Average & 0.73  & \textbf{0.78} \\
\hline
\end{tabular}%
\end{table}

\section{Conclusion and Limitations}
\label{sec:conclusion}

In this work, we proposed ViReSkill, a lifelong learning framework for robotic manipulation that combines vision-based replanning with skill accumulation and reuse. Our approach enables robust and efficient robot control by identifying and correcting failure cases using visual feedback, while storing successful plans as reusable skills to minimize reliance on model inference. Experiments on LIBERO, RLBench, and a physical UR5 robot demonstrated that ViReSkill outperforms existing baselines in both success rate and adaptability, confirming its practical potential in real-world environments.

Although ViReSkill supports replanning, its current execution remains open-loop with respect to perception: once a trajectory is generated, it is followed without real-time sensory feedback. While the system can replan between executions, it does not respond to observations during execution. We plan to extend the framework to closed-loop control by introducing Visual-Language Actions (VLA) as skills that generate actions dynamically based on image inputs at each timestep. Moreover, as the skill memory expands, retrieval cost and management efficiency may become bottlenecks. To address this, scalable skill management techniques such as skill merging, compression, and fast retrieval mechanisms will be essential.

By tackling these challenges, we aim to develop ViReSkill into a more general and sustainable learning framework that can adapt across diverse environments and robotic platforms.

\bibliographystyle{ieeetr}  
\bibliography{example}       


\end{document}